%% file: main.tex
\documentclass[runningheads]{llncs}
\usepackage[T1]{fontenc}
\usepackage{graphicx}
\usepackage{booktabs}
\usepackage[misc]{ifsym}
\usepackage{tikz}
\usetikzlibrary{decorations.pathreplacing}
\usetikzlibrary{positioning}
\newcommand{\corr}{(\Letter)}

\usepackage{amsmath, amssymb}
\usepackage{algorithm}
\usepackage{algorithmic}
\usepackage{xcolor}
\usepackage{colortbl}
\usepackage{booktabs}
\usepackage{multirow}
\usepackage{array}
\usepackage{adjustbox}
\usepackage{subcaption}
\usepackage{float}
\usepackage{wrapfig}

\usepackage[font=small, labelfont=bf]{caption}

\newtheorem{assumption}[theorem]{Assumption}

\begin{document}

\title{KANITE: Kolmogorov–Arnold Networks for ITE estimation}

\titlerunning{Kolmogorov–Arnold Networks for ITE estimation}

\author{Eshan Mehendale\thanks{This work was carried out during an internship at Sony Research India.}  \and
Abhinav Thorat \corr \and
Ravi Kolla \and 
Niranjan Pedanekar }
\authorrunning{Eshan Mehendale et al.}
\institute{Sony Research India, \\ 
\email{\{eshan.mehendale,
abhinav.thorat \corr, 
ravi.kolla, 
niranjan.pedanekar\}@sony.com}
}

\maketitle              

\begin{abstract}
We introduce \textbf{KANITE}, a framework leveraging Kolmogorov– Arnold Networks (KANs) for Individual Treatment Effect (ITE) estimation under multiple treatments setting in causal inference. By utilizing KAN's unique abilities to learn univariate activation functions as opposed to learning linear weights by Multi-Layer Perceptrons (MLPs), we improve the estimates of ITEs. The KANITE framework comprises two key architectures: 1.Integral Probability Metric (IPM) architecture: This employs an IPM loss in a specialized manner to effectively align towards ITE estimation across multiple treatments. 2. Entropy Balancing (EB) architecture: This uses weights for samples that are learned by optimizing entropy subject to balancing the covariates across treatment groups. Extensive evaluations on benchmark datasets demonstrate that KANITE outperforms state-of-the-art algorithms in both \textbf{PEHE} (\textbf{P}recision in the \textbf{E}stimation of \textbf{H}eterogeneous \textbf{E}ffects) and \textbf{ATE} (\textbf{A}verage \textbf{T}reatment \textbf{E}ffect) error metrics. Our experiments highlight the advantages of KANITE in achieving improved causal estimates, emphasizing the potential of KANs to advance causal inference methodologies across diverse application areas.

\end{abstract}

\input{introduction}

\input{literature-survey}

\input{problem-formulation}

\input{proposed-model}

\input{experiments}

\input{conclusion}

\bibliographystyle{splncs04}
\bibliography{kanite-bibliography}

\end{document}

%% file: introduction.tex
\section{Introduction}
\label{sec:introduction}
In causal inference, the estimation of Individual Treatment Effects (ITEs) is a foundational problem, as it is crucial for understanding the impact of a treatment on an individual user and personalizing treatments. In observational studies, the estimation of ITEs becomes particularly challenging due to the presence of confounders—variables that affect both the treatment and the outcome. For example, imagine a store that offers a discount on a high-end coffee machine only during periods of high customer volume, such as busy weekend hours. An analyst notices that customers who receive the discount are less likely to complete their purchase and concludes that the discount is ineffective. However, a hidden confounder—queue length—may be influencing both the likelihood of receiving the discount (since it is only offered during high-traffic times) and the decision to abandon the purchase (due to long wait times). In this case, queue length distorts the observed relationship between the discount and purchasing behavior. Consequently, it is essential to mitigate the bias introduced by such confounders in order to clearly isolate and estimate the treatment's effect on the outcome. ITE estimation is widely recognized to have applications across a broad range of domains, including, but not limited to, healthcare~\cite{stukel2007analysis}, education~\cite{hong2005effects}, e-commerce~\cite{chan2010evaluating}, entertainment~\cite{yu2022causal} and social sciences~\cite{jordan2012juvenile}. Given its importance, a wide range of algorithms has been developed to address this challenge, each adopting different modeling strategies and assumptions.

These approaches span from classical methods like propensity score matching to more recent advances in representation learning and deep neural networks. However, many of these approaches face trade-offs in flexibility, interpretability, and generalization. This motivates the need for more expressive and structured models such as the Kolmogorov–Arnold Network (KAN), which offers a promising framework for capturing complex causal relationships with greater clarity and adaptability.

In the year 2024, Kolmogorov-Arnold Networks (KANs) have been introduced as a promising alternative to Multi Layer Perceptrons (MLPs), also known as fully connected feedforward neural networks, offering the advantage of improved accuracy, interpretability and reduced model complexity~\cite{liu2024kan}. Although both MLPs and KANs feature fully connected structures, the key difference lies in their learning mechanisms. KANs learn univariate activation functions at each edge of network, whereas MLPs learn linear weights along all edges. Further, KANs are inspired by the Kolmogorov-Arnold representation theorem~\cite{kolmogorov1957representations},~\cite{braun2009constructive} whereas MLPs are motivated by the universal approximation theorem~\cite{hornik1989multilayer}. Shortly after their inception, KANs were rapidly integrated into various algorithmic frameworks, where they replaced MLPs and demonstrated notable performance improvements.
To that end, we direct the reader to the following references for a deeper understanding of KAN applications: transformer architectures~\cite{yang2024kolmogorov}, federated learning~\cite{zeydan2024f}, online reinforcement learning~\cite{kich2024kolmogorov}, autoencoders~\cite{moradi2024kolmogorov}, convolutional neural networks~\cite{bodner2024convolutional}, and graph neural networks~\cite{kiamari2024gkan}.

Although KANs have been applied in various domains, as mentioned above, their potential in the context of ITE estimation remains unexplored. To the best of our knowledge, this is the first study to investigate and propose algorithms that leverage KANs for ITE estimation in the multiple treatment setting. Given that mitigating confounding bias is critical for accurate ITE estimation, we aim to enhance this by utilizing KANs to better capture complex treatment and outcome relationships. Furthermore, since confounding bias becomes even more profound in the case of multiple treatments, we address this challenge by combining KANs with a loss function formulated using either Integral Probability Metrics (IPM) or the Entropy Balancing (EB) method. Additionally, we investigate the effect of KAN parameters such as grid size and spline degree on ITE estimation performance.

In the following we outline the salient contribution of our work. 
\begin{itemize}
    \item To the best of our knowledge, it is the first work that studies and incorporates KANs into the ITE estimation including the multiple treatment setting.
    
    \item We propose the KANITE framework for ITE estimation, which employs shared representation learning with representation loss formulated using either the IPM or EB method. Our KANITE framework comprises three distinct algorithms, inspired by~\cite{shalit2017estimating}, leveraging KANs as its fundamental building blocks. 
         
    \item To achieve improved covariate balancing across all treatments, we extend the entropy balancing method~\cite{zeng2020double} (originally developed for binary treatment settings) using Lagrangian duality theory to handle multiple treatments, and propose an algorithm that integrates both KANs and entropy balancing loss. 
    
    \item Through extensive numerical evaluations, we demonstrate the superior performance of KANITE against baselines on various binary and multiple treatments benchmark datasets such as IHDP, NEWS-2/4/8/16, ACIC-16 and Twins. 
    
    \item We also provide a detailed analysis of the impact of various KAN parameters such as grid size and the degree of splines used in the univariate activation functions for ITE estimates.
     
\end{itemize}

We structure the rest of the paper as follows. The next section reviews related work and highlights key differences. Section~\ref{sec:problem-formulation} provides the technical details underlying the problem formulation. Section~\ref{sec:proposed-model} presents our proposed models and their technicalities in detail. Section~\ref{sec:experiments} covers the baselines and compares them with KANITE on the PEHE and ATE error metrics. Finally, Section~\ref{sec:conclusion} concludes the paper and suggests future research directions.

%% file: literature-survey.tex
\section{Literature survey}
\label{sec:literature-survey}
This section briefly reviews relevant literature and contrasts it with our contributions. To the best of our knowledge, this is the first work to explore the utilization of KANs in ITE estimation. Therefore, we review the literature on ITE estimation and KANs separately.

ITE estimation has been extensively studied in the literature; thus, we restrict our discussion to a few notable works. In \cite{shalit2017estimating}, \cite{schwab2018perfect}, and \cite{yoon2018ganite}, the authors address an ITE estimation setup similar to ours and propose efficient algorithms based on MLPs—hence, these works have been chosen as baselines in our work. Additionally, in~\cite{guo2020learning} and \cite{thorat2023estimation}, the authors consider ITE estimation in a network setting, where users are assumed to be connected through a network. They propose algorithms that leverage additional user network information to obtain improved ITE estimates.  
A few other works~\cite{harada2021graphite}, \cite{kaddour2021causal},  \cite{nilforoshan2023zero} and~\cite{thorat2024see} incorporate auxiliary treatment information rather than treating treatments categorically, demonstrating methods to achieve improved ITE estimates. Moreover, leveraging treatment information inherently endows algorithms with zero-shot capabilities, enabling them to predict the outcomes for novel treatments that were \textit{not} encountered during training. It is important to note that these approaches\textemdash network-based ITE estimation and the use of auxiliary treatment information\textemdash are distinct from the setup considered in this study. 

In~\cite{liu2024kan}, the authors introduce KANs and demonstrate their advantages over MLPs in terms of accuracy, model complexity, and interpretability\textemdash both theoretically and empirically. Since then, KANs have been incorporated in various areas of research, consistently demonstrating their potential benefits. In~\cite{kiamari2024gkan}, the authors propose two methods to integrate KAN layers into graph convolutional networks and empirically evaluate these architectures using a semi-supervised graph learning task using the Cora dataset. In~\cite{yang2024kolmogorov}, a KAN-based transformer architecture is proposed that employs rational functions over splines in the KAN layers to enhance model expressiveness and performance. Meanwhile, \cite{zeydan2024f} introduces a KAN-based federated learning approach that outperforms its MLP counterparts on classification tasks. In~\cite{kich2024kolmogorov} the use of KANs in the proximal policy optimization algorithm is explored, demonstrating benefits in terms of model complexity. The authors in~\cite{moradi2024kolmogorov} investigate the efficiency of KANs for data representation through autoencoders, while KAN-based convolutional neural networks are proposed and evaluated on the Fashion-MNIST dataset in~\cite{bodner2024convolutional}, showcasing advantages over their MLP counterparts. Additionally, KANs have been employed in physics-informed deep learning frameworks to improve the modeling of physical systems. In~\cite{wang2024kolmogorov}, the authors introduce Kolmogorov–Arnold-Informed Neural Networks (KINN), which leverage KANs in place of traditional MLPs to solve both forward and inverse problems governed by differential equations. In a separate line of work~\cite{patra2024physics}, the authors propose Physics-Informed Kolmogorov–Arnold Networks (PIKAN), which incorporate Efficient-KAN and WAV-KAN architectures and demonstrate their superior performance compared to conventional physics-informed neural networks based on MLPs.

%% file: problem-formulation.tex
\section{Problem Formulation}
\label{sec:problem-formulation}
In this section, we present the mathematical formulation of the problem considered in this work. We adopt the Rubin-Neyman~\cite{rubin2005causal} potential outcomes framework to introduce the problem. For clarity, we define the following notation. Let $N$ and $K$ denote the number of users (samples) and treatments respectively. We use $\mathbf{x}_i$ and $t_i$ to denote the covariates and assigned treatment of user-$i$ respectively. Furthermore, Let $Y_t^i$ denote the potential outcome for user-$i$ when treatment-$t$ is given. For brevity, when the context is clear, we may omit the user index in the notation. We assume that the following standard causal inference assumptions from~\cite{rubin2005causal} hold.

\begin{assumption}[Unconfoundedness]
\label{assumption:uncofoundedness}
Under this assumption, the potential outcomes, $Y_t$'s, are independent of the treatment assignment, $t$, conditioned on the user covariates, $\mathbf{x}.$ Mathematically, stated as:
\begin{equation*}
%\label{eq:unconfoundedness}
( Y_1, Y_2, \cdots, Y_K ) \perp t \mid \mathbf{x}. 
\end{equation*}
In other words, this assumption ensures that all confounders, covariates that are affecting both $Y_t$ and $t$, are observed and accounted in $\mathbf{x}.$ 
\end{assumption}

\begin{assumption}[Positivity]
\label{assumption:positivity} 
It ensures that each user has a positive probability of receiving any of the available treatments. Mathematically it is given as:
\begin{equation*}
\mathbb{P}(t_i = t) > 0 \quad \forall 1 \leq i \leq N, \, \, 1 \leq t \leq K. 
\end{equation*}
\end{assumption}

\begin{assumption}[Stable Unit Treatment Value Assumption (SUTVA)]
\label{assumption:sutva} 
It implies that the potential outcomes of a user are solely dependent on their received treatments and independent of the assigned treatments of other users. 
\end{assumption}

With the help of the above, let us define the ITE and ATE of treatment-$a$ with respect to $b$ for a user with covariates, $\mathbf{x}_i$, denoted by $\tau_{a, b}(\mathbf{x}_i)$ and $\textrm{ATE}_{a, b}$ respectively, as:
\begin{align}
\tau_{a, b}(\mathbf{x}_i) & = \mathbb{E} \left[ Y_a^i - Y_b^i \mid \mathbf{x} = \mathbf{x}_i \right] \label{eq:ITE} \\
\textrm{ATE}_{a, b} & = \mathbb{E} \left[ Y_a - Y_b \right]. \label{eq:ATE}
\end{align}
We now introduce the problem as follows. Given $N$ samples $\lbrace  \mathbf{x}_i, t_i, Y_{t_i}^i \rbrace_{i=1}^N,$ our goal is to estimate ITEs of all users and ATEs across all pairs of treatments. We use the existing error metrics~\cite{thorat2023estimation} for this problem, such as $\epsilon_{\text{PEHE}}$ and $\epsilon_{\text{ATE}}$, to quantify the performance of a model, as defined below:
\begin{align}
\epsilon_{\text{PEHE}}  &= \frac{1}{{K \choose 2}} \sum\limits_{a=1}^{K} \sum\limits_{b=1}^{a-1}\left[ \frac{1}{N} \sum\limits_{i=1}^N (\hat{\tau}^{a, b}(\mathbf{x}_i) - \tau^{a, b}(\mathbf{x}_i))^2 \right] \label{eq:epsilon-pehe-formula}  \\
\epsilon_{\text{ATE}}  &= \frac{1}{{K \choose 2}} \sum\limits_{a=1}^{K} \sum\limits_{b=1}^{a-1}\left[ \left| \frac{1}{N} \sum\limits_{i=1}^N \hat{\tau}^{a, b}(\mathbf{x}_i) - \frac{1}{N}\sum\limits_{i=1}^N \tau^{a, b}(\mathbf{x}_i) \right| \right], \label{eq:epsilon-ate-formula}  
\end{align}
where $\hat{\tau}\left( \cdot \right)$ represents the estimated ITEs produced by the model.

%% file: proposed-model.tex
\section{Proposed Model}
\label{sec:proposed-model}
In this section, we present our proposed framework KANITE (Kolmogorov-Arnold Networks for Individual Treatment Effect estimation), that leverages KANs for causal inference, specifically for estimating ITEs. KANITE utilizes the functional decomposition properties of KANs, which decompose complex functions into sum of univariate functions. This decomposition enables KANITE to capture the causal effect of a treatment while accounting for confounding variables that influence both treatment assignment and outcomes. 
KANITE’s ability to approximate any continuous function allows it to adapt to diverse data distributions, establishing it as a flexible and effective framework for causal inference. 
It operates under the standard assumptions of causal inference stated in Assumption~\ref{assumption:uncofoundedness}, \ref{assumption:positivity} and \ref{assumption:sutva}. We provide a brief overview of KAN preliminaries below, which is a crucial part of the KANITE framework.
\begin{figure}
\label{fig:KAN}
    \centering
   \includegraphics[scale=0.25]{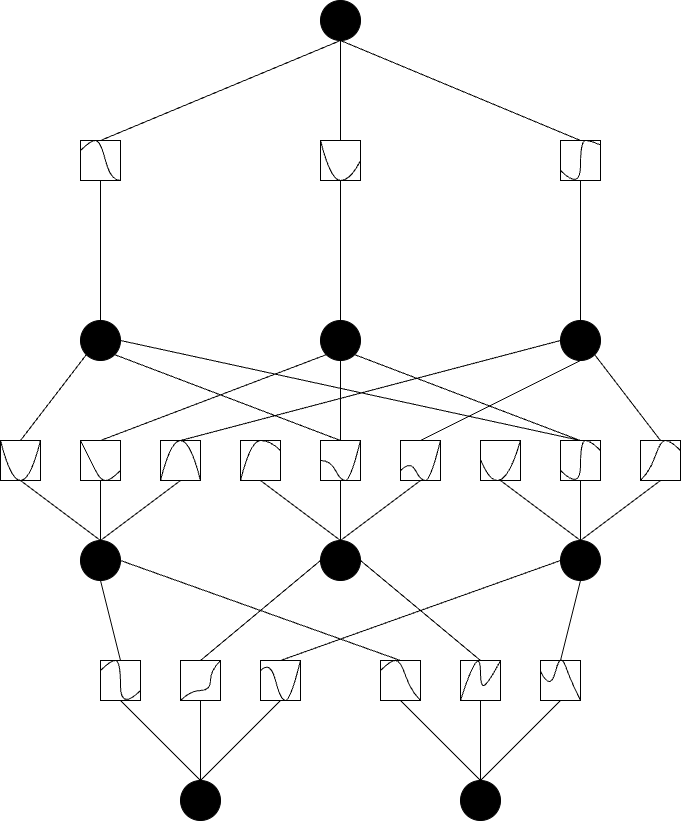}
    \caption{KAN: Kolmogorov-Arnold Networks~\cite{liu2024kan}} 
    \label{fig:KAN}
\end{figure}
\subsection{KAN Preliminaries}
KANs have recently emerged as a significant advancement in a wide range of tasks that rely on predictive algorithms at their core. While their effectiveness in supervised learning has been well-documented, to the best of our knowledge, no work has yet explored their application to causal inference. The foundation of KANs lies in the Kolmogorov-Arnold Representation Theorem~\cite{Braun2009OnAC}, which states that any smooth function $f: [0,1]^n \to \mathbb{R}$ can be expressed as:
\begin{equation}
\label{eq:KAN theorem}
f(x_1, \dots, x_n) = \sum_{q=1}^{2n+1} \Phi_q \left( \sum_{p=1}^{n} \phi_{q,p}(x_p) \right),
\end{equation}
where $\phi_{q,p}: [0, 1] \rightarrow \mathbb{R}$ and $\Phi_q: \mathbb{R} \rightarrow \mathbb{R}.$ This formulation demonstrates that any smooth multivariate function can be fundamentally decomposed into a sum of univariate functions, making the composition purely additive.
This theorem serves as the inspiration for the KAN architecture, originally proposed for supervised learning tasks. In such tasks, the goal is to model a function $f$ based on input-output pairs $\{(\mathbf{x}_i, y_i)\}$, such that $y_i \approx f ( \mathbf{x}_i)$.

The KAN architecture as illustrated in Figure \ref{fig:KAN} is designed such that all learnable functions are univariate, with each parameterized using basis functions, such as a B-spline, to enhance the model's flexibility. Liu et al.~\cite{liu2024kan} introduced the KANs, initially proposing a two-layer model where learnable activation functions are placed on the edges, with aggregation achieved through summation at the nodes. However, this simple design had limitations in approximating complex functions. To address these shortcomings, the authors extended the approach within the same work by introducing multiple layers and increasing both the breadth and depth of the network, thereby enhancing its ability to approximate more complex functions. Mathematically, a typical $l^{\text{th}}$ KAN layer, suitable for deeper architectures, with $n_l$ inputs $(x^l_1, x^l_2, \cdots, x^l_{n_l})$ and $n_{l+1}$ outputs $(x^{l+1}_1, x^{l+1}_2, \cdots, x^{l+1}_{n_{l+1}})$ is defined as follows: 
\begin{equation}
\label{eq:KANlayer}
\begin{bmatrix}
x^{l+1}_1 \\
x^{l+1}_2 \\
\vdots \\
x^{l+1}_{n_{l+1}}
\end{bmatrix} =
\begin{bmatrix}
\phi_{1, 1} & \phi_{1, 2} &  \cdots & \phi_{1, n_l}\\
\phi_{2, 1} & \phi_{2, 2} &  \cdots & \phi_{2, n_l} \\
\vdots & \vdots & \vdots & \vdots \\
\phi_{n_{l+1}, 1} & \phi_{n_{l+1}, 2} &  \cdots & \phi_{n_{l+1}, n_l} 
\end{bmatrix}
\cdot
\begin{bmatrix}
x^l_1 \\
x^l_2 \\
\vdots \\
x^l_{n_l}
\end{bmatrix},
\end{equation}
where each $\phi_{q,p}$ $\forall$ $p \in \lbrace 1, 2, \dots, n_l \rbrace $ and $q \in \lbrace 1, 2, \dots, n_{l+1} \rbrace$ is a trainable univariate function with adjustable parameters. This structure allows the original two-layer Kolmogorov-Arnold representation, given in~\eqref{eq:KAN theorem}, to be extended into a more robust, deeper architecture capable of handling increasingly complex tasks. With the help of the above, we now proceed to explain KANITE framework in detail in the following subsection.
\begin{algorithm}
\caption{KANITE Training}
\begin{algorithmic}
\STATE \textbf{Input:} Observational data: $ \mathcal{D} = \{ \left( \mathbf{x}_i, t_i, Y_{t_i}^i \right) \}_{i=1}^n \sim \mathcal{D}_{\text{train}},\mathcal{D}_{\text{val}}$, and hyper parameters $\alpha \geq 0$ and $\beta \geq 0.$
\STATE \textbf{Output:} An outcome prediction model: $f ( \Psi,\Pi ),$ where $\Pi = (\Pi_1, \Pi_2, \cdots, \Pi_K )$ 
\end{algorithmic}
\begin{algorithmic}[1]
\setcounter{ALC@line}{0}
\STATE Initialize parameters: $\Psi: \mathtt{KAN},$ $\Pi_i: \mathtt{KAN}$  $\forall i \in \lbrace 1, 2, \cdots, K \rbrace$
\WHILE{{\textit{not converged}}}
    \STATE Sample a mini-batch  \\ $\mathcal{B} = \{ (\mathbf{x}_{i_o}, Y^{i_o}_{ t_{i_o}}) \}_{o=1}^B \subset \mathcal{D}_{\text{train}} $
    \STATE Mini-batch approximation of Regression Loss \\ 
    $\mathcal{L}_1 = \frac{1}{B} \sum\limits_{o=1}^{B} (\hat{Y}_{t_{i_o}} - Y_{t_{i_o}})^2 $
    \STATE Mini-batch approximation of the Representation Loss \\ $\mathcal{L}_2 = \frac{1}{\binom{K}{2}} \sum\limits_{a=1}^{K} \sum\limits_{b=1}^{a-1} 
\mathrm{Representation Loss}\left( \Psi_{t=a},  \Psi_{t=b} \right)$
    \STATE Update Functions: \\
    $f(\Psi,\Pi) \leftarrow f ( \Psi,\Pi) - \lambda.\nabla( f ({\Psi,\Pi}))$
    \STATE Minimize $\alpha \cdot \mathcal{L}_1 + \beta \cdot \mathcal{L}_2$
    using SGD
\ENDWHILE
\end{algorithmic}
\end{algorithm}

\subsection{KANITE Architecture}
Our proposed KANITE framework addresses the task of ITE estimation for multiple treatments by utilizing KANs as the backbone of its architecture. Figure~\ref{fig:kanite architecture} illustrates the details of the KANITE framework, explained through the following three key steps. 
\begin{enumerate}
    \item [A.] \textit{Balanced Representation of Covariates:} First, KANITE aims to learn a balanced covariate representation by replacing the conventional MLPs with the KANs, shown as Representation Network in Figure~\ref{fig:kanite architecture}, enabling the model to
    learn latent representations of covariates balanced across all treatment groups.
    
    \item [B.] \textit{Treatment Head Networks:} It consists of dedicated treatment head networks, where each treatment is modeled through a separate representation using KANs, allowing greater flexibility to capture the underlying distribution of treatment outcomes.

    \item [C.] \textit{Representation loss:} Three different representation losses have been considered in the proposed set of algorithms under KANITE. First and second losses are Maximum Mean Discrepancy (MMD) and Wasserstein, based on the Integral Probability Metric (IPM), and the third one utilizes Entropy Balancing (EB) method~\cite{zeng2020double} to learn weights that minimize the Jensen-Shannon divergence, asymptotically, between all pairs of treatment groups. These three losses result into three different algorithms named KANITE-MMD, KANITE-Wass, KANITE-EB for ITE estimation.
\end{enumerate}

\input{kanite-architecture-diagram}
This approach enables us to utilize the KANs for the ITE estimation task in an effective manner while simultaneously learning covariate and treatment representations that improve upon state-of-the-art (SOTA) ITE estimation algorithms. A detailed explanation of the KANITE framework is as follows.

\subsubsection{KANs for learning balanced covariate representation}
In earlier ITE estimation literature, representation learning for covariates has demonstrated a significant improvement~\cite{shalit2017estimating}. 
In the KANITE architecture, we utilize a KAN layer setup for achieving a balanced representation that caters to multiple treatment scenarios. KAN layers, as defined in Equation \ref{eq:KANlayer}, enable the architecture to learn low-dimensional symbolic representations of covariates, which help separate treatment-related signals from confounding influences, thereby mitigating confounding bias~\cite{shalit2017estimating}. To learn representations for covariates, $\mathbf{x} \in \mathcal{X}$, we employ KAN layers setup with learnable activation functions consist of B-Splines that learns a balanced representation function, \(\Psi : \mathcal{X} \rightarrow \mathbb{R}^d\), in a lower-dimensional latent space.

In KANITE, a deep representation network is constructed by stacking multiple KAN layers one after the other to form a hierarchical model for representation learning. Let $L$ denote the total number of KAN layers. For each layer $l \in \lbrace 0, 1, \cdots, L-1 \rbrace,$ let $n_l$	represent the total number of neurons in layer~$l.$ Let $\Psi^l(\mathbf{x}) = \left( \Psi^{l}_1(\mathbf{x}), \Psi^{l}_2(\mathbf{x}), \cdots, \Psi^{l}_{n_{l}}(\mathbf{x})  \right)$ denote the representation after the $(l-1)^{\text{st}}$ layer, with the input defined as \(\Psi^{0}(\mathbf{x}) = \mathbf{x}\). In contrast to standard MLPs, KANs do not learn independent weight or bias parameters; instead, each layer aggregates the outputs of learnable univariate activation functions. The transformation at layer $l \in \lbrace0,1,...,L-1\rbrace,$ denoted as $\Psi^{l+1}(\mathbf{x}) = \left( \Psi^{l+1}_1(\mathbf{x}), \Psi^{l+1}_2(\mathbf{x}), \cdots, \Psi^{l+1}_{n_{l+1}}(\mathbf{x})  \right),$ is defined in a compositional form analogous to the Kolmogorov--Arnold representation as follows:
\begin{equation}
    \label{eq:deepkan1}
\Psi^{l+1}_i(\mathbf{x})=\sum_{j=1}^{n_{l}} \phi^{l}_{i, j}\left(\Psi^{l}_j(\mathbf{x})\right), \quad \forall i \in \lbrace 1, 2, \cdots, n_{l+1} \rbrace.
\end{equation}
Using the recursion, we can write the above as:
\begin{equation}
    \label{eq:kan-layer-recursive-equation}
    \Psi^{l+1}(\mathbf{x}) = \left( \Psi^{l} \circ \Psi^{l-1} \circ \cdots \circ \Psi^{1} \circ \Psi^{0} \right) \mathbf{x}.
\end{equation}
This recursive formulation enables the deeper architecture to capture complex, non-linear interactions among covariates, progressively refining the balanced representation and further mitigating confounding bias for improved treatment effect estimation.

\subsubsection{Treatment Head Networks}
The KANITE framework leverages a balanced covariate representation learned from the representation network to drive the treatment head networks for enhanced treatment-specific ITE estimation. As depicted in the KANITE architecture, we deploy a distinct treatment head network for each unique treatment category. Deep KAN layers, given in  Equation~\ref{eq:kan-layer-recursive-equation}, are trained to learn a symbolic representation function that is specific to treatments. We denote these treatment head networks by \(\Pi_t\) for \(t \in \{1, 2, 3, \cdots, K\}\), where \(K\) is the total number of unique treatments. Consider a user with covariates \(\mathbf{x}\) and the assigned treatment as $t$, the treatment head network \(\Pi_t\) with $M$ number of layers is defined as:
\begin{equation}
    \label{eq:pi-equation}
\Pi^{M+1}_{t}(\mathbf{x}) = \left( \Pi^{M}_{t} \circ \Pi^{M-1}_{t} \circ \cdots \circ \Pi^{1}_{t} \circ \Pi^{0}_{t} \right) \Psi(\mathbf{x}),
\end{equation}
where $\Psi(\mathbf{x})$ is the balanced representation of covariates learned from the representation network. 
Furthermore, we leverage network sparsification in KANs to reduce the impact of redundant activation functions, which acts as a form of regularization and improves ITE estimates.

\subsubsection{Representation Loss}
As mentioned in the previous subsections, learning a balanced covariates representation across all treatments plays a crucial role in KANITE. Hence, we employ three variations of the representation loss in KANITE based on IPM and Entropy Balancing, resulting into three different algorithms, in addition to the standard Mean Square Error (MSE) loss on the observed factual data. Note that, MSE loss is defined as: $\mathcal{L}_1 = \frac{1}{n} \sum_{i=1}^{n} (\hat{Y}_{i, t} - Y_i)^2.$ In the below we provide more details of the representation loss variants.   

\begin{itemize}
    \item[1.] \textbf{IPM based representation loss:} \\IPMs have shown promising results in achieving balanced representations for ITE estimation, as demonstrated in \cite{shalit2017estimating} and \cite{johansson2016learning}. In KANITE, we leverage two popular IPM-based loss functions—the Maximum Mean Discrepancy (MMD) and the Wasserstein loss, to effectively capture distributional differences between treatment subgroups. MMD is particularly useful because it compares higher-order moments between distributions, minimizing subtle discrepancies in the feature space, while the Wasserstein metric provides a robust measure of distance even when distributions have limited overlap. 
    For our multiple-treatment setup, we use the average pairwise IPM loss from~\cite{thorat2023estimation} to learn a balanced representation across all treatment group combinations. The mathematical formulation is provided below:
    \begin{equation}
        \label{eq:ipm-loss}
\mathcal{L}_2 = \frac{1}{\binom{K}{2}} \sum_{a=0}^{K-1} \sum_{b=0}^{a-1} \texttt{IPM} \left( \lbrace \Psi \rbrace_{t=a}, \lbrace \Psi \rbrace_{t=b} \right),
    \end{equation}
     where $\texttt{IPM}()$ can be either MMD or Wasserstein, leading to the respective algorithms KANITE-MMD and KANITE-Wass. 

\item [2.]\textbf{Entropy Balancing (EB) based representation loss:}

In \cite{zeng2020double}, a doubly robust representation learning approach is proposed for ITE estimation in the binary treatment setting. It uses Entropy Balancing (EB) to learn weights that, in the limit, minimize the Jensen-Shannon divergence between treated and control covariates distributions. In this work, we extend this methodology to the multiple-treatment setting to balance covariate distributions across all treatment groups, as given below.  

Let $m$ be the number of covariates. Let $t$ and $s$ denote the indicies of treatments i.e., $t, s \in \lbrace 1, 2, \cdots, K \rbrace.$ Entropy balancing optimization problem for the multiple-treatment setting to balance covariates distributions is given as:
\begin{align}
& \hspace{2.5cm} \mathbf{w}^{\text{EB}} = \arg\max_{\mathbf{w}} \left\{ -\sum_{i=1}^{N} w_i \log w_i \right\}, \nonumber\\[1ex]
& \text{s.t.}
\begin{cases}
\text{(i)} \sum\limits_{T_i = t} w_i\,\Psi(x_{ji}) = \sum\limits_{T_i = s} w_i\,\Psi(x_{ji}), \forall\, j \in \{1,2,\ldots,m\} \text{ and } t < s,\\[1ex]
\text{(ii)} \sum\limits_{T_i = t} w_i = 1, \forall\, t \in \{1, 2, \cdots, K\}, \forall w_i > 0.
\end{cases}
\label{eq:original-eb-for-multiple-treatments}
\end{align}
Note that, constraint (i) ensures that the weighted sum of the shared representations of the covariates is balanced across all pairs of treatment combinations. Then, the representation loss in this case is given as 
\begin{equation}
    \label{eq:EB-representation-loss}
    \mathcal{L}_2 = \sum\limits_{i=1}^N w_i^{\text{EB}}(\Psi) \log ( w_i^{\text{EB}} (\Psi) ). 
\end{equation}
We solve the optimization problem in \eqref{eq:original-eb-for-multiple-treatments} by formulating its dual problem using Lagrangian duality theory~\cite{boyd2004convex}.  To that end, let us define the following: for $t < s$, $\lambda_{t,s} = [\lambda_{t,s,1}, \lambda_{t,s,2}, \ldots, \lambda_{t,s,m}] \in \mathbb{R}^m
$ and set $\lambda_{t,s} = -\lambda_{s,t} \quad \text{for } t > s.$
By constructing the Lagrangian function and applying the Karush-Kuhn-Tucker (KKT) conditions we get the following dual problem of \eqref{eq:original-eb-for-multiple-treatments}: 
\begin{equation}
\label{eq:dual-eb-for-multiple-treatments}
\min_{\lambda_{t,s}} \sum_{t=1}^{k} \log \left( \sum_{T_i = t} \exp \left( - \sum_{s \neq t} \langle \lambda_{t,s}, \Psi_i \rangle \right) \right),
\end{equation}
where $\Psi_i = [\Psi_{i,1}, \Psi_{i,2}, \ldots, \Psi_{i,m}] \in \mathbb{R}^m. $
Using the above dual formulation, we now provide a closed-form solution for equation \eqref{eq:original-eb-for-multiple-treatments}. Suppose \(T_i = t\); then, the weight for sample-$i$, $w_i$, is given by:
\begin{equation}
\label{eq:weight-equation}
w_i = \frac{\exp \left( - \sum_{s \neq t} \langle \lambda_{t,s}, \Psi_i \rangle \right)}{\sum_{T_i = t} \exp \left( - \sum_{s \neq t} \langle \lambda_{t,s}, \Psi_i \rangle \right)}.
\end{equation}
This formulation provides a principled approach to deriving weights using the EB method that, in the limit, minimize the JSD. We refer to the algorithm that employs the EB-based representation loss as KANITE-EB.
\end{itemize}
Note that the final loss function of the KANITE framework is a weighted sum of the standard MSE and the chosen representation loss (MMD, Wasserstein, or EB-based), as shown below.
\begin{equation*}
    \mathcal{L} = \alpha \cdot \mathcal{L}_1 + \beta \cdot \mathcal{L}_2 \quad \text{for some} \, \alpha, \beta >0.
\end{equation*}

%% file: kanite-architecture-diagram.tex
%Architecture Diagram
% Architecture Diagram
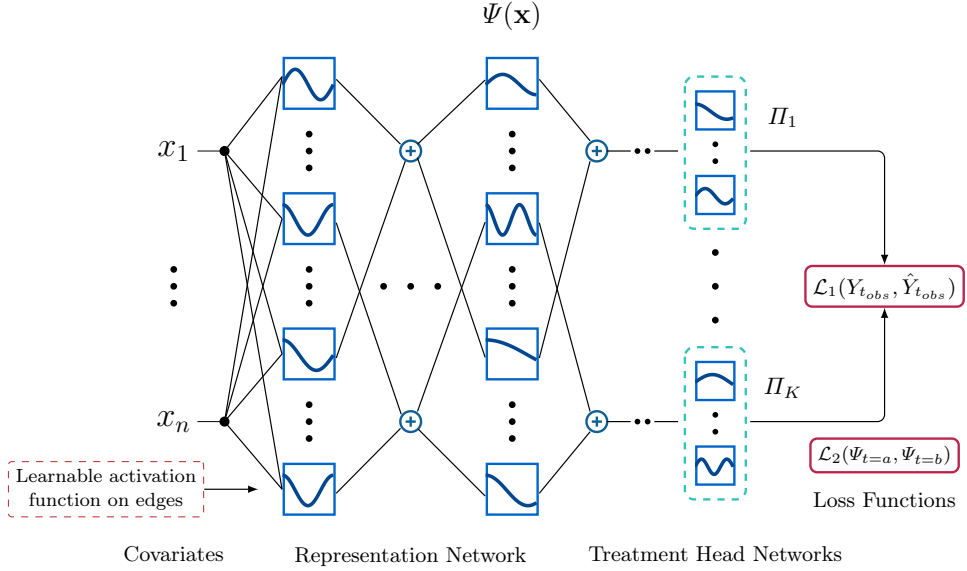
\begin{figure}
    \centering
    \scalebox{0.92}{
    \begin{tikzpicture}
        % Define colors
        \definecolor{kanblock}{HTML}{0066CC}
        \definecolor{kanwave}{HTML}{00478F}
        \definecolor{pluso}{HTML}{105F93}
        \definecolor{hnhl}{HTML}{32C3BB}
        \definecolor{info}{HTML}{BA2926}
        \definecolor{loss}{HTML}{BB254B}
        % Define a reusable command with a unique node name parameter:
        % #1 = node name, #2 = node position, #3 = wave function,
        % #4 = domain for the plot, #5 = wave color.
        \newcommand{\WaveBlock}[6]{%
  \pgfmathsetmacro{\rectSize}{0.75*#6}%
  % Draw the rectangular (square) node with scaled size
  \node[draw=kanblock, line width=1pt, minimum width=\rectSize cm, minimum height=\rectSize cm, align=center] (#1) at #2 {};%
  % Clip the drawing to the node's interior and scale the coordinate system
  \begin{scope}
    \clip (#1.south west) rectangle (#1.north east);
    \begin{scope}[shift={(#1.south west)}, xscale=\rectSize, yscale=\rectSize]
      \draw[#5, line width=1.5pt, domain=#4, samples=50]
        plot (\x, {#3});
    \end{scope}
  \end{scope}
}

        \newcommand{\PlusoNode}[3]{%
    \pgfmathsetmacro{\plusoffset}{0.25*#3}%
    % Draw the circular node with the specified size
    \node[draw=pluso, circle, line width=1pt, minimum size={#3 cm}] (#1) at #2 {};%
    % Draw the horizontal line of the plus sign
    \draw[pluso, line width=1pt]
      ([shift={(-\plusoffset,0)}]#1.center) -- ([shift={(\plusoffset,0)}]#1.center);%
    % Draw the vertical line of the plus sign
    \draw[pluso, line width=1pt]
      ([shift={(0,-\plusoffset)}]#1.center) -- ([shift={(0,\plusoffset)}]#1.center);%
  }
        
        %Representation Network First column
        \WaveBlock{square1}{(0,0)}{0.5 + 0.3*sin(2*pi*\x r)}{0:1}{kanwave}{1};

        \node[circle, fill=black, inner sep=1pt] at (0,-0.75) {};
        \node[circle, fill=black, inner sep=1pt] at (0,-1) {};
        \node[circle, fill=black, inner sep=1pt] at (0,-1.25) {};
        
        \WaveBlock{square2}{(0,-2)}{0.5 + 0.3*cos(2*pi*\x r)}{0:1}{kanwave}{1};
        
        \node[circle, fill=black, inner sep=1pt] at (0,-2.75) {};
        \node[circle, fill=black, inner sep=1pt] at (0,-3) {};
        \node[circle, fill=black, inner sep=1pt] at (0,-3.25) {};
        
        \WaveBlock{square2}{(0,-4)}{0.5 + 0.3*cos(1.5*pi*\x r)}{0:1}{kanwave}{1};

        \node[circle, fill=black, inner sep=1pt] at (0,-4.75) {};
        \node[circle, fill=black, inner sep=1pt] at (0,-5) {};
        \node[circle, fill=black, inner sep=1pt] at (0,-5.25) {};

        \WaveBlock{square2}{(0,-6)}{0.5 + 0.3*cos(2*pi*\x r)}{0:1}{kanwave}{1};

        %\PlusoNode{nodeA}{(1.5,-1)}{0.5}

        %\PlusoNode{nodeA}{(1.5,-5)}{0.5}
        %% Second representation KAN columns
        \node at (1.75,1) [font=\fontsize{12}{14}\selectfont\bfseries] {$\Psi(\mathbf{x})$};
        \draw[line width=0.5pt,decorate, decoration={brace, amplitude=8pt}, yshift=0pt]
    (-0.45,0.45) -- (3.45,0.45);
        \WaveBlock{square1}{(3,0)}{0.5 + 0.2*sin(1.5*pi*\x r)
}{0:1}{kanwave}{1};
        
        \node[circle, fill=black, inner sep=1pt] at (3,-0.75) {};
        \node[circle, fill=black, inner sep=1pt] at (3,-1) {};
        \node[circle, fill=black, inner sep=1pt] at (3,-1.25) {};
        
        \WaveBlock{square2}{(3,-2)}{0.5 + 0.3*cos(3*pi*\x r)}{0:1}{kanwave}{1}
        
        \node[circle, fill=black, inner sep=1pt] at (3,-2.75) {};
        \node[circle, fill=black, inner sep=1pt] at (3,-3) {};
        \node[circle, fill=black, inner sep=1pt] at (3,-3.25) {};
        
        \WaveBlock{square2}{(3,-4)}{0.5 + 0.3*cos(0.6*pi*\x r)}{0:1}{kanwave}{1}

        \node[circle, fill=black, inner sep=1pt] at (3,-4.75) {};
        \node[circle, fill=black, inner sep=1pt] at (3,-5) {};
        \node[circle, fill=black, inner sep=1pt] at (3,-5.25) {};

        \WaveBlock{square2}{(3,-6)}{0.5 + 0.3*cos(1.2*pi*\x r)}{0:1}{kanwave}{1};

        %Covariates
         \node at (-2,-6.95) [font=\fontsize{10}{12}] {Covariates};
         \node at (-2,-1) [font=\fontsize{14}{16}] {$x_1$};
         \node[circle, fill=black, inner sep=1.5pt] at (-1.25,-1) {};
         \node at (-2,-5) [font=\fontsize{14}{16}] {$x_n$};
         \node[circle, fill=black, inner sep=1.5pt] at (-1.25,-5) {};
        \node[circle, fill=black, inner sep=1pt] at (-2,-2.75) {};
        \node[circle, fill=black, inner sep=1pt] at (-2,-3) {};
        \node[circle, fill=black, inner sep=1pt] at (-2,-3.25) {};
        % covariates to dot lines
        \draw (-1.65,-1) -- (-1.25,-1);
        \draw (-1.65,-5) -- (-1.25,-5);
        \draw[line width=0.5pt] (-1.25,-1) --(-0.4,0.1);
        \draw[line width=0.5pt] (-1.25,-1) -- (-0.4,-2);
        \draw[line width=0.5pt] (-1.25,-1) -- (-0.4,-4);
        \draw[line width=0.5pt] (-1.25,-1) -- (-0.4,-6);
        %xn node
        \draw[line width=0.5pt] (-1.25,-5) -- (-0.4,0.1);
        \draw[line width=0.5pt] (-1.25,-5) -- (-0.4,-2);
        \draw[line width=0.5pt] (-1.25,-5) -- (-0.4,-4);
        \draw[line width=0.5pt] (-1.25,-5) -- (-0.4,-6);

        %representaton layer 2 connecting nodes
        \PlusoNode{nodeA}{(1.5,-1)}{0.3};
         %\PlusoNode{nodeA}{(1.5,-3)}{0.3};
          \PlusoNode{nodeA}{(1.5,-5)}{0.30};
        
        %% Lines for first KAN layer to additive node
        %\node[circle, fill=black, inner sep=1pt] at (1.35,-1) {};
         \draw[line width=0.5pt] (0.4,0.06) --(1.35,-0.9);
        \draw[line width=0.5pt] (0.4,-6.06) --(1.35,-5.1);
        \draw[line width=0.5pt] (0.4,-2.06) --(1.36,-4.89);
         \draw[line width=0.5pt] (0.4,-4.06) --(1.42,-1.15);

        %% Lines for Additive Node to second KAN layer
        \draw[line width=0.5pt] (2.6,0.05)--(1.6,-0.9);
         \draw[line width=0.5pt] (1.6,-4.85) --(2.6,-2.05);
         \draw[line width=0.5pt] (1.6,-1.1) --(2.6,-4.05);
         \draw[line width=0.5pt] (1.6125,-5.1125) --(2.6,-6.1);
        % \draw[line width=0.35pt] (1.58,-3.1) --(2.6,-6.05);
        % \draw[line width=0.35pt] (1.55,-4.85) --(2.6,-4.05);

        \node at (1.5,-7) [font=\fontsize{10}{12}] {Representation Network};

        %Head networks block top block
        \WaveBlock{square1}{(6,-0.4)}{0.5 + 0.2*cos(1.2*pi*\x r) }{0:1}{kanwave}{0.75};
        
        \node[circle, fill=black, inner sep=0.85pt] at (6,-0.9) {};
        \node[circle, fill=black, inner sep=0.85pt] at (6,-1.15) {};
        \WaveBlock{square1}{(6,-1.65)}{0.5 + 0.2*sin(1.9*pi*\x r)}{0:1}{kanwave}{0.75};

        %\node at (6,0.025) [font=\fontsize{10}{11}] {$\Pi_0$};
        %\node at (6,-4) [font=\fontsize{10}{11}] {$\Pi_k$};
        
        %dots between
        \node[circle, fill=black, inner sep=1pt] at (6,-2.5) {};
        \node[circle, fill=black, inner sep=1pt] at (6,-3) {};
        \node[circle, fill=black, inner sep=1pt] at (6,-3.5) {};
        
        %head network bottom block
        \WaveBlock{square1}{(6,-4.4)}{0.5 + 0.2*sin(1.1*pi*\x r) }{0:1}{kanwave}{0.75};
        
        \node[circle, fill=black, inner sep=0.85pt] at (6,-4.9) {};
        \node[circle, fill=black, inner sep=0.85pt] at (6,-5.15) {};
        \WaveBlock{square1}{(6,-5.65)}{0.5 + 0.2*sin(3*pi*\x r)}{0:1}{kanwave}{0.75};

        \node at (6,-6.95) [font=\fontsize{10}{12}] {Treatment Head Networks};

        \PlusoNode{nodeA}{(4.25,-1)}{0.3};
          \PlusoNode{nodeA}{(4.25,-5)}{0.30};

        % representation KAN layer 2 to plus nodes
        \draw[line width=0.5pt] (3.4,0.06) --(4.15,-0.85);
        \draw[line width=0.5pt] (3.4,-6.06) --(4.15,-5.15);
        \draw[line width=0.5pt] (3.4,-2.06) --(4.15,-4.85);
        \draw[line width=0.5pt] (3.4,-4.06) --(4.15,-1.15);

        %rectangle around tp head netowork
        \draw[dashed, rounded corners, draw=hnhl, line width=1pt] (5.55,0.1) rectangle (6.45,-2.15);

        \draw[dashed, rounded corners, draw=hnhl, line width=1pt] (5.55,-3.9) rectangle (6.45,-6.15);

        \node at (7,-0.5) [font=\fontsize{10}{12}\selectfont\bfseries] {$\Pi_1$};
        \node at (7,-4.5) [font=\fontsize{10}{12}\selectfont\bfseries] {$\Pi_K$};

        %psi(x) plus nodes lines dots to pi 0 heads
        \draw[line width=0.5pt] (4.4,-1)--(4.75,-1);
        \node[circle, fill=black, inner sep=0.85pt] at (4.85,-1) {};
        \node[circle, fill=black, inner sep=0.85pt] at (5,-1) {};
        \draw[line width=0.5pt] (5.10,-1)--(5.45,-1);

        %psi(x) plus nodes lines dots to pi k heads
        \draw[line width=0.5pt] (4.4,-5)--(4.75,-5);
        \node[circle, fill=black, inner sep=0.85pt] at (4.85,-5) {};
        \node[circle, fill=black, inner sep=0.85pt] at (5,-5) {};
        \draw[line width=0.5pt] (5.10,-5)--(5.45,-5);
        
        %hoizontal dots between KAN layers
        \node[circle, fill=black, inner sep=1pt] at (1.1,-3) {};
        \node[circle, fill=black, inner sep=1pt] at (1.5,-3) {};
        \node[circle, fill=black, inner sep=1pt] at (1.9,-3) {};

        %learnable activation fucntion
        \node[draw=info, rectangle, dashed, align=center, rounded corners=0.5mm, font=\fontsize{8}{10}\selectfont] at (-3,-6)
    {Learnable activation \\ function on edges};
\draw[->,>=latex,line width=0.5pt] (-1.55,-6)--(-0.75,-6);

    %loss functions
    \node[draw=loss,line width=0.35mm,          % Thicker outline
      rounded corners=1mm, rectangle, align=center, font=\fontsize{9}{10}\selectfont] at (8.5,-3)
    {$\mathcal{L}_1(Y_{t_{obs}}, \hat{Y}_{t_{obs}})$};

    \node[draw=loss,line width=0.35mm,          % Thicker outline
      rounded corners=1mm, rectangle, align=center, font=\fontsize{8}{9}\selectfont] at (8.5,-5.5)
    {$\mathcal{L}_2(\Psi_{t=a},\Psi_{t=b})$};
    \node at (8.5,-6.15) [font=\fontsize{10}{12}] {Loss Functions};

    % lines connecting plus node and pi blocks
    %\draw[line width=0.5pt] (4.40,-1.1) --(5.45,-5);
    %\draw[line width=0.5pt] (4.37,-4.87) --(5.45,-1);

    %loss lines 
    % Define start and end points
    \node (A) at (6.4,-1) {};
    \node (B) at (8.5,-2.8) {};

    % Draw the line with a soft downward curve and arrow
    \draw[->,>=latex, line width=0.5pt, rounded corners=3pt]
        (A) -- ++(2.1, 0) -- (B);
        %pik to l1
        % Define start and end points
    \node (A) at  (6.4,-5) {};
    \node (B) at (8.5,-3.2) {};
     \draw[->,>=latex, line width=0.5pt, rounded corners=3pt]
        (A) -- ++(2.1, 0) -- (B);
    \end{tikzpicture}
    }
    \caption{KANITE Architecture}
    \label{fig:kanite architecture}
\end{figure}

%% file: experiments.tex
\section{Experiments}
\label{sec:experiments}
In this section, we present a detailed numerical analysis of KANITE's performance on several standard benchmark datasets \textemdash IHDP~\cite{shalit2017estimating}, NEWS~\cite{schwab2018perfect}, TWINS~\cite{almond2005costs}, and ACIC-16~\cite{dorie2019automated} \textemdash compared to baselines. We first evaluate KANITE against baselines using the metrics $\epsilon_{\text{PEHE}}$ and $\epsilon_{\text{ATE}}$, as defined in Equations~\eqref{eq:epsilon-pehe-formula} and~\eqref{eq:epsilon-ate-formula}. Next, we examine its convergence and parameter efficiency relative to the baselines. Finally, we analyze the impact of hyperparameters, such as grid size and spline degree in activation functions of KAN layers, on ITE estimation. 

\subsection{Baselines}
We compare KANITE with various baseline architectures designed for ITE estimation in both binary and multiple-treatment settings. For the multiple-treatment evaluation, we use the NEWS-4, NEWS-8, and NEWS-16 semi-synthetic datasets from~\cite{schwab2018perfect}, while for binary treatment setting, we consider the IHDP, TWINS, ACIC-16, and NEWS-2 datasets. The models we benchmark against include TarNet, CFRNet-Wass, and CFRNet-MMD~\cite{shalit2017estimating} which utilize IPM as the representation loss.  We also introduce a baseline called CFRNet-EB, which uses the Entropy Balancing loss, as given in Equation~\eqref{eq:EB-representation-loss}, in place of IPM within the CFRNet architecture.
For a fair comparison in the multiple-treatment setting, we also compare KANITE with Perfect Match \cite{schwab2018perfect}. Additionally, to benchmark against generative counterfactual predictive models, we evaluate GANITE \cite{yoon2018ganite}. Since KANITE operates in both binary and multiple treatment scenarios, we appropriately modify baselines developed for binary treatment setting, such as TarNet, CFRNet-Wass, CFRNet-MMD and CFRNet-EB to ensure a fair comparison across both treatment setups. 

\begin{table}
\centering
\caption{Performance comparison of KANITE vs baselines on $\epsilon_{\text{PEHE}}$ metric across various binary treatment setting datasets}
\label{tab:pehe_comparison_binary_treatments}
\begin{tabular}{l@{\hspace{7pt}}c@{\hspace{7pt}}c@{\hspace{7pt}}c@{\hspace{7pt}}c}
\hline
\textbf{Method/Dataset} & \textbf{IHDP} & \textbf{NEWS-2} & \textbf{TWINS} & \textbf{ACIC-16} \\
\hline
TarNet & 2.33 $\pm$ 2.71 & 23.90 $\pm$ 8.75 & \textbf{0.32 $\pm$ 0.00} & 2.41 $\pm$ 0.91\\
CFRNet-Wass & 1.50 $\pm$ 1.76 & 23.85 $\pm$ 6.24 & \textbf{0.32 $\pm$ 0.00} & 2.58 $\pm$ 1.05\\
CFRNet-MMD & 1.50 $\pm$ 1.73 & 23.14 $\pm$ 7.10  & \textbf{0.32 $\pm$ 0.00} & 2.42 $\pm$ 0.88\\
CFRNET-EB &  1.22 $\pm$ 1.32  & 21.25 $\pm$ 5.33 & 0.43 $\pm$ 0.20 & 2.89 $\pm$ 1.44\\
PerfectMatch & 1.56 $\pm$ 1.71 & 23.18 $\pm$ 8.13 & \textbf{0.32 $\pm$ 0.00} & 2.48 $\pm$ 0.89\\
GANITE & 7.91 $\pm$ 7.47 & 23.22 $\pm$ 8.38 & 0.35 $\pm$ 0.07 & 5.24 $\pm$ 1.38\\
\rowcolor{gray!20}
KANITE-Wass & \textbf{1.08 $\pm$ 1.39} & 20.78 $\pm$ 3.59 & \textbf{0.32 $\pm$ 0.00} & \textbf{1.58 $\pm$ 1.09} \\
\rowcolor{gray!20}
KANITE-MMD & \textbf{1.08 $\pm$ 1.39} & 20.78 $\pm$ 3.61 & \textbf{0.32 $\pm$ 0.00} & \textbf{1.58 $\pm$ 1.09} \\
\rowcolor{gray!20}
KANITE-EB & \textbf{1.08 $\pm$ 1.39} &  \textbf{20.32 $\pm$ 2.82} & \textbf{0.32 $\pm$ 0.00} & \textbf{1.58 $\pm$ 1.09} \\
\hline
\end{tabular}
\end{table}

\begin{table}
\centering
\caption{Performance comparison of KANITE vs baselines on $\epsilon_{\text{ATE}}$ metric across various binary treatment setting datasets}
\label{tab:ate_comparison_binary_treatments}
\begin{tabular}{l@{\hspace{7pt}}c@{\hspace{7pt}}c@{\hspace{7pt}}c@{\hspace{7pt}}c}
\hline
\textbf{Method/Dataset} & \textbf{IHDP} & \textbf{NEWS-2} & \textbf{TWINS} & \textbf{ACIC-16} \\
\hline
TarNet & 0.63 $\pm$ 0.83 & 11.85 $\pm$ 11.50 & 0.02 $\pm$ 0.01 & 0.30 $\pm$ 0.16\\
CFRNet-Wass & 0.24 $\pm$ 0.25 & 11.61 $\pm$ 9.48 & 0.02 $\pm$ 0.01 & 0.54 $\pm$ 0.20\\
CFRNet-MMD & 0.24 $\pm$ 0.24 & 10.85 $\pm$ 9.73  & 0.01 $\pm$ 0.01 & 0.37 $\pm$ 0.32\\
CFRNET-EB & 0.29 $\pm$ 0.34  & 7.71 $\pm$ 7.46 & 0.22 $\pm$ 0.28 & 0.37 $\pm$ 0.19\\
PerfectMatch & 0.25 $\pm$ 0.25 & 10.34 $\pm$ 10.64 & 0.03 $\pm$ 0.01 & 0.39 $\pm$ 0.29\\
GANITE & 4.40 $\pm$ 1.33 & 11.28 $\pm$ 10.80 & 0.35 $\pm$ 0.07 & 3.61 $\pm$ 1.07\\
\rowcolor{gray!20}
KANITE-Wass & \textbf{0.15 $\pm$ 0.13} & 7.03 $\pm$ 5.43 & \textbf{0.01 $\pm$ 0.00} & \textbf{0.18 $\pm$ 0.13} \\
\rowcolor{gray!20}
KANITE-MMD & \textbf{0.15 $\pm$ 0.13} & 7.02 $\pm$ 5.48 & \textbf{0.01 $\pm$ 0.00} & 0.19 $\pm$ 0.14 \\
\rowcolor{gray!20}
KANITE-EB & \textbf{0.15 $\pm$ 0.13} &  \textbf{6.38 $\pm$ 4.49} & \textbf{0.01 $\pm$ 0.00} & \textbf{0.18 $\pm$ 0.13} \\
\hline
\end{tabular}
\end{table}

\begin{table}
\centering
\caption{Performance comparison of KANITE vs baselines on $\epsilon_{\text{PEHE}}$ metric across various multiple treatment setting datasets}
\label{tab:pehe_comparison_multiple_treatments}
\begin{tabular}{l@{\hspace{7pt}}c@{\hspace{7pt}}c@{\hspace{7pt}}c@{\hspace{7pt}}c}
\hline
\textbf{Method/Dataset} & \textbf{NEWS-4} & \textbf{NEWS-8} & \textbf{NEWS-16} \\
\hline
TarNet & 24.09 $\pm$ 4.07 & 24.85 $\pm$ 6.73 & 25.06 $\pm$ 2.96 \\
CFRNet-Wass & 24.98 $\pm$ 4.57 & 22.70 $\pm$ 3.39 & 22.60 $\pm$ 1.75 \\
CFRNet-MMD & 24.05 $\pm$ 4.56 & 23.17 $\pm$ 3.32  & 22.81 $\pm$ 1.63 \\
CFRNET-EB & 21.71 $\pm$ 2.63  & 22.53 $\pm$ 3.13 & 22.33 $\pm$ 1.69 \\
PerfectMatch & 23.90 $\pm$ 4.60 & 23.41 $\pm$ 4.20 & 23.33 $\pm$ 1.68 \\
GANITE & 23.77 $\pm$ 4.10 & 24.10 $\pm$ 3.33 & 22.85 $\pm$ 1.62 \\
\rowcolor{gray!20}
KANITE-Wass & \textbf{21.48 $\pm$ 2.27} & \textbf{22.48 $\pm$ 3.31} & 22.20 $\pm$ 1.57 \\
\rowcolor{gray!20}
KANITE-MMD & 21.53 $\pm$ 2.31 & 22.58 $\pm$ 3.37 & \textbf{22.19 $\pm$ 1.57} \\
\rowcolor{gray!20}
KANITE-EB & 21.52 $\pm$ 2.30 &  22.62 $\pm$ 3.38 & 22.20 $\pm$ 1.58 \\
\hline
\end{tabular}
\end{table}

\begin{table}
\centering
\caption{Performance comparison of KANITE vs baselines on $\epsilon_{\text{ATE}}$ metric across various multiple treatment setting datasets}
\label{tab:ate_comparison_multiple_treatments}
\begin{tabular}{l@{\hspace{7pt}}c@{\hspace{7pt}}c@{\hspace{7pt}}c@{\hspace{7pt}}c}
\hline
\textbf{Method/Dataset} & \textbf{NEWS-4} & \textbf{NEWS-8} & \textbf{NEWS-16} \\
\hline
TarNet & 11.87 $\pm$ 5.07 & 10.91 $\pm$ 3.49 & 12.47 $\pm$ 3.01 \\
CFRNet-Wass & 13.33 $\pm$ 5.56 & 9.08 $\pm$ 3.55 & 9.08 $\pm$ 1.96 \\
CFRNet-MMD & 11.43 $\pm$ 5.64 & 9.98 $\pm$ 3.37  & 9.09 $\pm$ 1.88 \\
CFRNET-EB & 8.26 $\pm$ 3.29  & 	9.03 $\pm$ 3.10 & 9.08  $\pm$ 1.92 \\
PerfectMatch & 11.43 $\pm$ 5.71 & 9.62 $\pm$ 3.63 & \textbf{8.85 $\pm$ 1.99} \\
GANITE & 11.65 $\pm$ 5.03 & 11.74 $\pm$ 3.50 & 10.54 $\pm$ 1.77\\
\rowcolor{gray!20}
KANITE-Wass & \textbf{7.92 $\pm$ 2.97} & \textbf{8.91 $\pm$ 3.09} & 9.49 $\pm$ 1.84 \\
\rowcolor{gray!20}
KANITE-MMD & 8.05 $\pm$ 2.95 & 9.24 $\pm$ 3.45 & 9.46 $\pm$ 1.84 \\
\rowcolor{gray!20}
KANITE-EB & 8.03 $\pm$ 2.93 &  9.30 $\pm$ 3.48 & 9.47 $\pm$ 1.85 \\
\hline
\end{tabular}
\end{table}

\begin{figure}
    \centering
    \begin{subfigure}[b]{0.48\textwidth}
        \includegraphics[width=\textwidth]{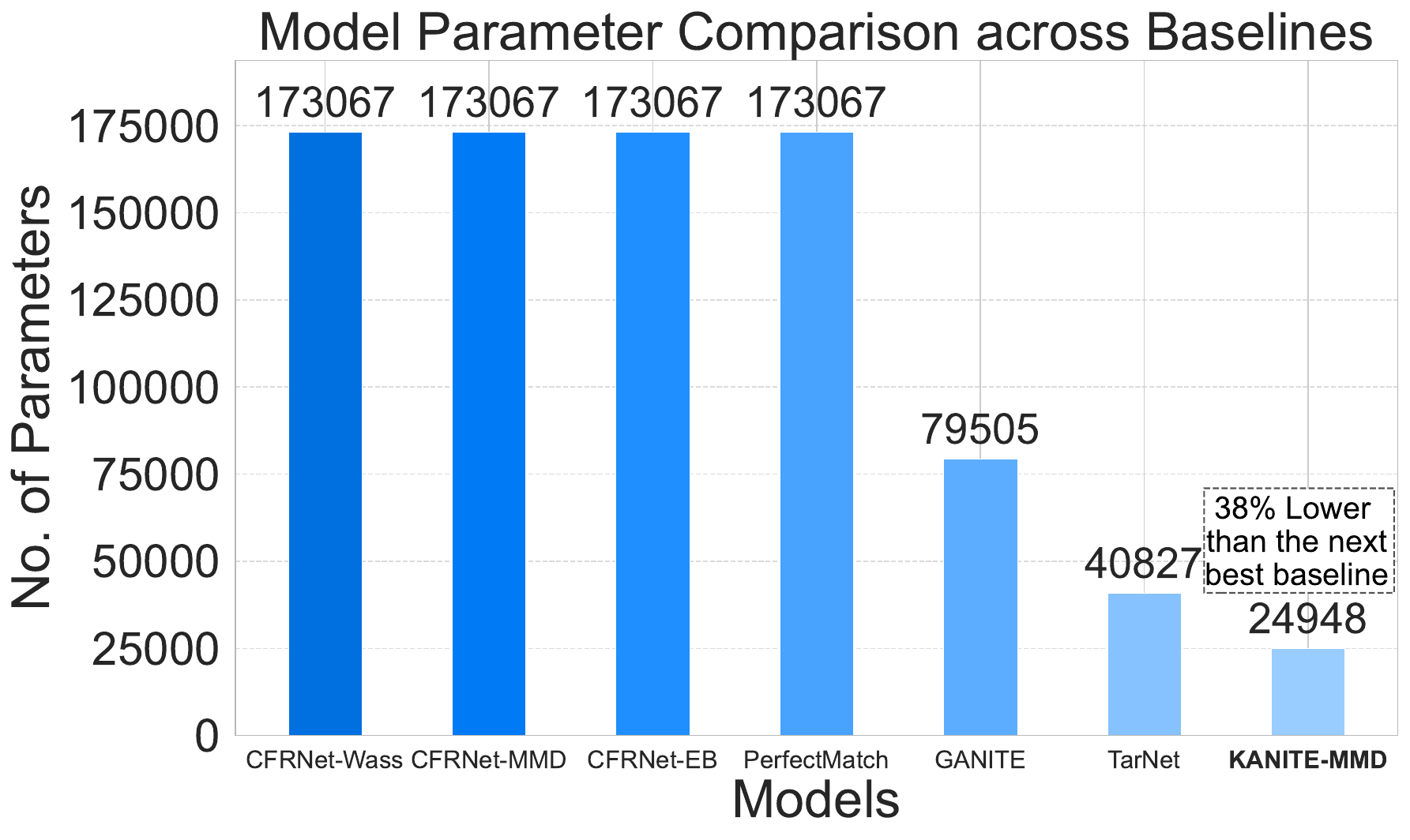}
        \caption{Comparison of model parameters}
        \label{fig:model-parm-comparison}
    \end{subfigure}
    \hfill
    \begin{subfigure}[b]{0.48\textwidth}
        \includegraphics[width=\textwidth]{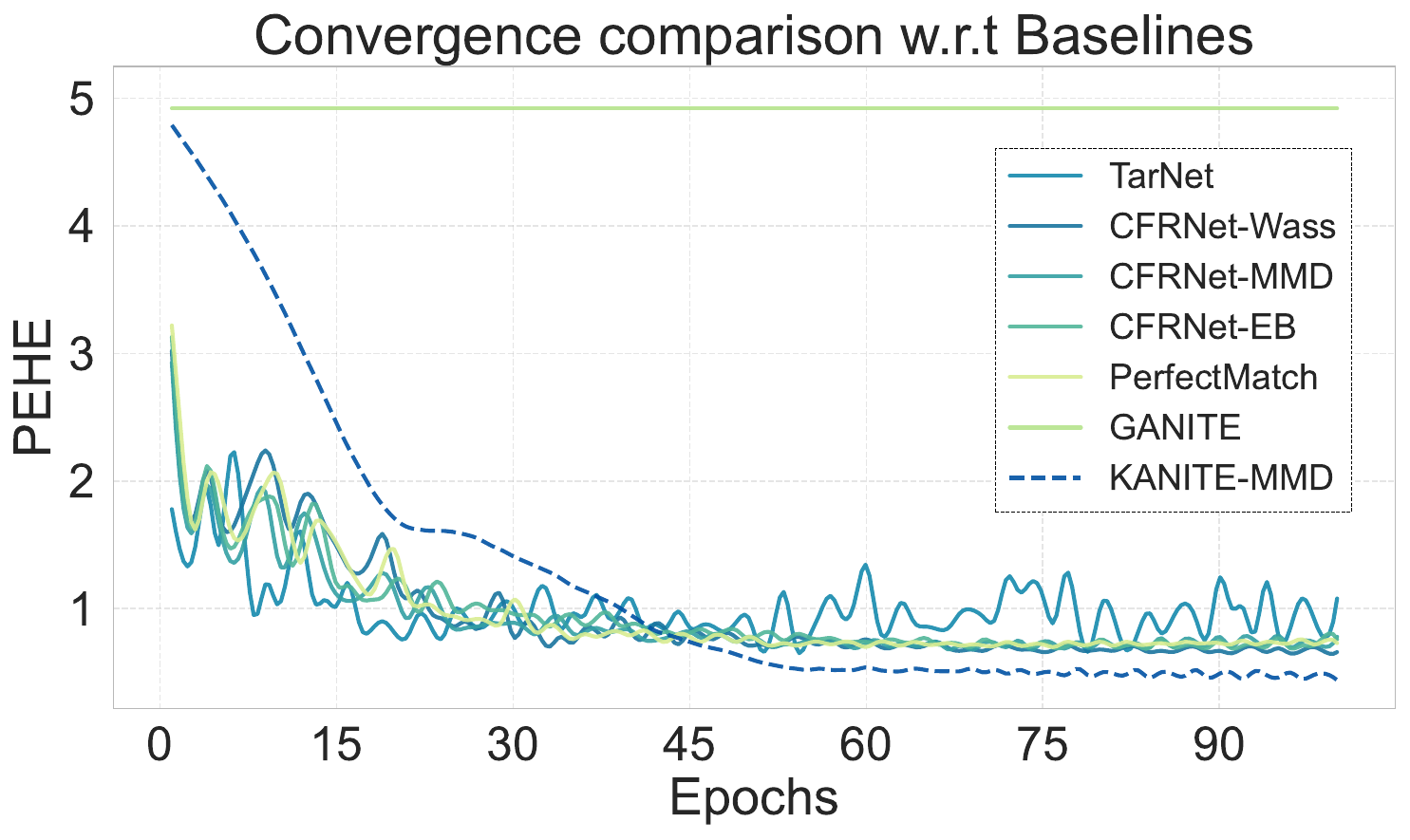}
        \caption{Comparison of model convergence}
        \label{fig:pehe-convergence}
    \end{subfigure}
    \captionsetup{font=normalsize}
    \caption{Comparison of model parameters and convergence across models}
    \label{fig:model-params-and-training-loss-convergence}
\end{figure}

\begin{figure}[!ht]
    \centering
    \begin{subfigure}[b]{0.48\textwidth}
        \includegraphics[width=\textwidth]{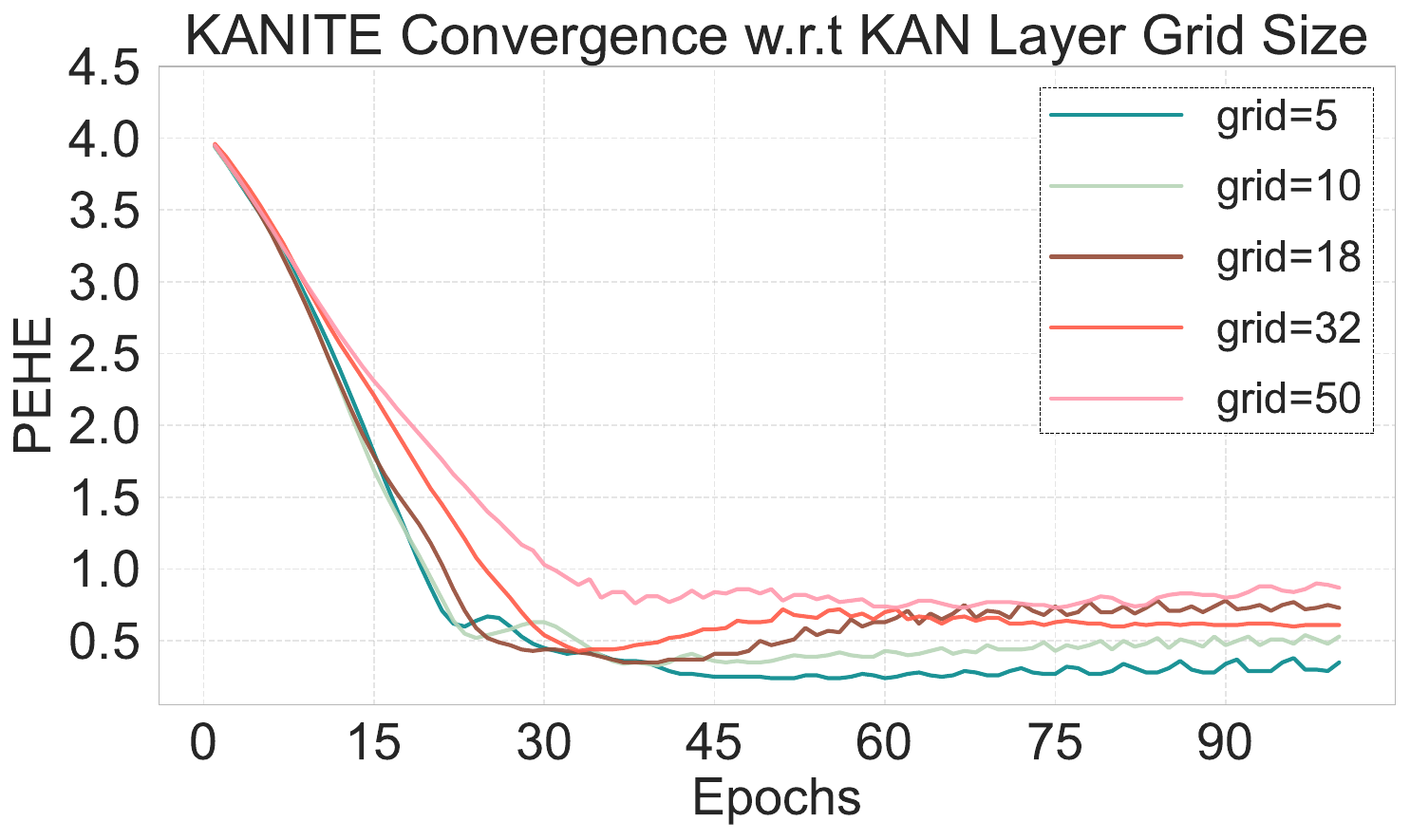}
        \caption{Affect of grid size}
        \label{fig:grid-size-comparison}
    \end{subfigure}
    \hfill
    \begin{subfigure}[b]{0.48\textwidth}
        \includegraphics[width=\textwidth]{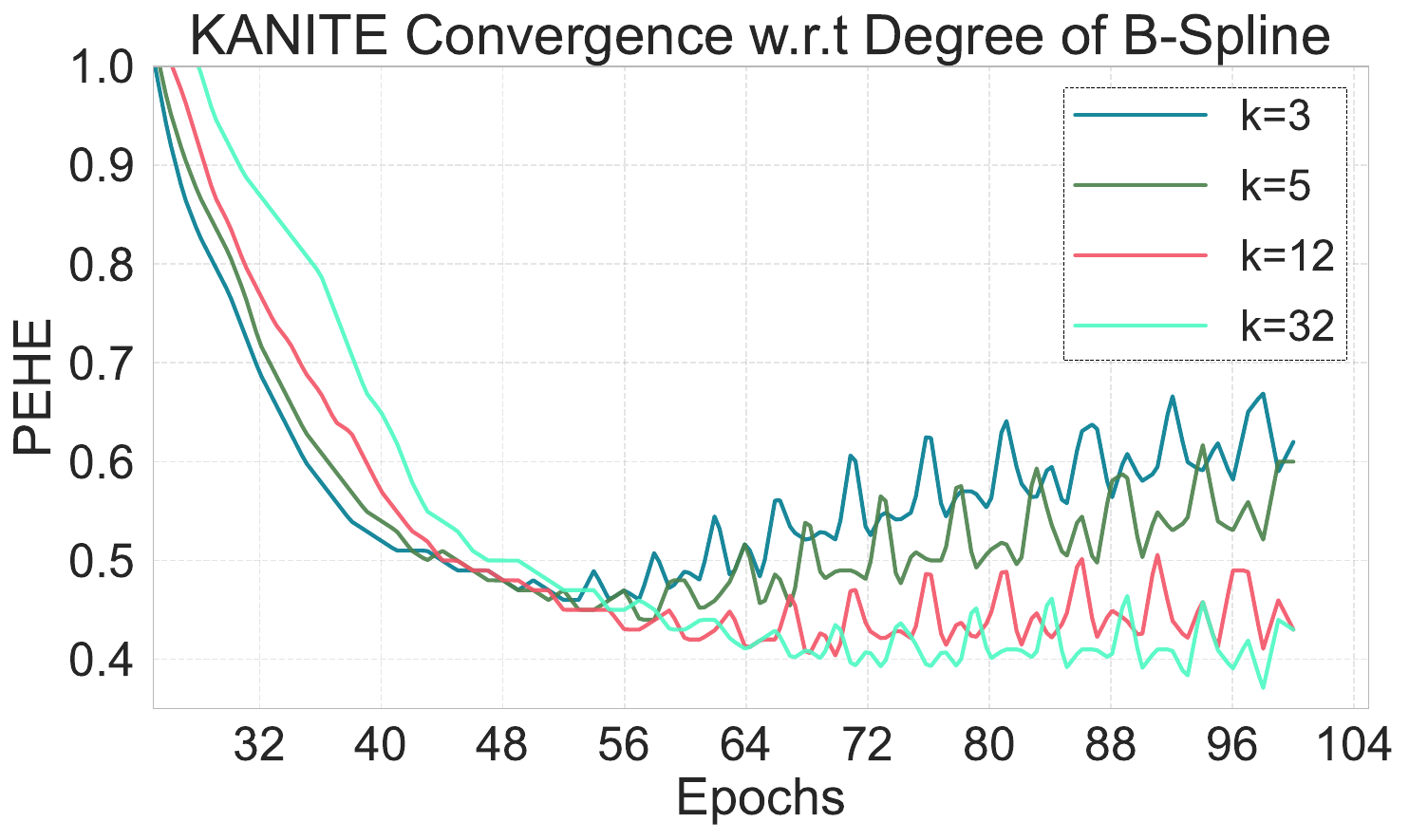}
        \caption{Affect of spline degree}
        \label{fig:spline-degree-convergence}
    \end{subfigure}
    \captionsetup{font=normalsize}
    \caption{Affect of grid size and spline degree considered in KANITE on ITE}
    \label{fig:grid-size-and-spline-degree-comparison}
\end{figure}

\subsection{KANITE: Performance Assessment}
We split the dataset into training, validation, and test sets in a 63:27:10 ratio. The results in all tables are computed on the full dataset after model training. We conducted 1000, 50, 10, and 10 iterations for the IHDP, NEWS, TWINS, and ACIC-16 datasets, respectively, and report the mean and standard deviation of these runs in the results tables. The best results in the tables are highlighted in bold. 

As mentioned earlier, the KANITE framework consists of three algorithms: KANITE-MMD, KANITE-Wass, and KANITE-EB, almost at least one of which outperforms all the baselines on both \(\epsilon_{\text{PEHE}}\) and \(\epsilon_{\text{ATE}}\) metrics in both binary and multiple treatment settings, as shown in Table~\ref{tab:pehe_comparison_binary_treatments}, \ref{tab:ate_comparison_binary_treatments}, \ref{tab:pehe_comparison_multiple_treatments}, and~\ref{tab:ate_comparison_multiple_treatments}. Note that Tables~\ref{tab:pehe_comparison_binary_treatments} and \ref{tab:ate_comparison_binary_treatments} present the \(\epsilon_{\text{PEHE}}\) and \(\epsilon_{\text{ATE}}\) metrics for all considered algorithms in the binary treatment setting, respectively. Similarly, Tables~\ref{tab:pehe_comparison_multiple_treatments} and~\ref{tab:ate_comparison_multiple_treatments} provide the corresponding results for the multiple-treatment setting. To perform a comprehensive performance assessment of KANITE, we evaluate its convergence and parameter efficiency compared to the baselines. Figure~\ref{fig:model-parm-comparison} compares the number of parameters in our proposed KANITE framework against all baselines. Notably, KANITE outperforms all baselines on both \(\epsilon_{\text{PEHE}}\) and \(\epsilon_{\text{ATE}}\) metrics while reducing model parameters by 38\% compared to the next best baseline. Figure~\ref{fig:pehe-convergence} shows that our proposed KANITE model, depicted in dotted line, converges faster than all baselines. Since all three KANITE variants exhibited similar behavior in terms of parameter count and convergence, we present only KANITE-MMD in Figure~\ref{fig:model-params-and-training-loss-convergence} to keep the figures uncluttered.

\subsection{KANITE: Hyperparameters study}
We now examine the impact of the B-Spline degree and grid size considered in KAN layers on model performance. As grid size and spline degree are direct proportional to the model complexity in terms of parameters we conduct the hyperparameter optimization on them and use the best parameters in the respective models. For example, Figure~\ref{fig:grid-size-and-spline-degree-comparison} shows the affect of grid size and spline degree on the ITE estimates for IHDP dataset. From Figure~\ref{fig:grid-size-and-spline-degree-comparison}, it can be observed that grid size of 5 and spline degree of 32 achieve the best performance on this iteration of the results.

%% file: conclusion.tex
\section{Conclusion}
\label{sec:conclusion}
In this study, we proposed KANITE, a state-of-the-art framework for ITE estimation that leverages shared representation learning using either IPM or Entropy Balancing. Unlike traditional MLP-based architectures, KANITE employs KANs as its backbone, enabling it to learn more accurate causal effect estimates. The framework introduces three algorithms—KANITE-MMD, KANITE-Wass, and KANITE-EB—each utilizing a different IPM or Entropy Balancing-based representation loss to ensure balanced covariate representations across treatment groups. Additionally, we derive a closed-form Entropy Balancing-based representation loss for the multiple-treatment setting using Lagrangian duality theory. Experimental results demonstrate that KANITE effectively handles multiple-treatment scenarios, outperforming all considered baselines on both the \(\epsilon_{\text{PEHE}}\) and \(\epsilon_{\text{ATE}}\) metrics. Furthermore, KANITE achieves superior parameter efficiency and faster convergence while maintaining strong counterfactual prediction capabilities.

For future work, we plan to further enhance KANITE to create a unified architecture that incorporates abilities of both IPM and Entropy Balancing for ITE estimation tasks. We plan to incorporate interpretability of KANs to understand causal effects estimation in a better manner. Our findings highlight the advantages of KANs in ITE estimation, paving the way for future research in related areas.  One promising direction is investigating the role of KANs in ITE estimation under networked settings, where users are interconnected through a network~\cite{thorat2023estimation}. Another avenue is examining the effectiveness of KANs in treatment dosage settings, where treatments are administered in fractional amounts between 0 and 1~\cite{schwab2020learning}. Additionally, it would be valuable to investigate how KANs can enhance causal effect estimation when treatment information is explicitly incorporated~\cite{harada2021graphite}.